\documentclass[10pt,twocolumn,letterpaper]{article}

\usepackage{iccv}
\usepackage{times}
\usepackage{epsfig}
\usepackage{graphicx}
\usepackage{amsmath}
\usepackage{amssymb}

\usepackage{booktabs}

\usepackage{algorithm}
\usepackage{algorithmic}

\usepackage[utf8]{inputenc} % allow utf-8 input
\usepackage[T1]{fontenc}    % use 8-bit T1 fonts
\usepackage{url}            % simple URL typesetting
\usepackage{booktabs}       % professional-quality tables
\usepackage{amsfonts}       % blackboard math symbols
\usepackage{nicefrac}       % compact symbols for 1/2, etc.
\usepackage{microtype}      % microtypography
\usepackage{xcolor}         % colors

\usepackage{graphicx}       % table width
\usepackage{amsmath}
\usepackage{amssymb}
\usepackage{multicol}
\usepackage{makecell}
\usepackage{multirow}
\usepackage{enumitem,kantlipsum}
\usepackage{wrapfig,lipsum,booktabs}

\newcommand{\mathbbm}[1]{\text{\usefont{U}{bbm}{m}{n}#1}}

\usepackage[pagebackref=true,breaklinks=true,letterpaper=true,colorlinks,bookmarks=false]{hyperref}

\iccvfinalcopy % *** Uncomment this line for the final submission

\def\iccvPaperID{5942} % *** Enter the ICCV Paper ID here

\ificcvfinal\fi

\begin{document}

%%%%%%%%% TITLE
% \title{Boosting Long-tailed Object Detection via \N{confidence}-guided Step-wise Learning}

\title{Boosting Long-tailed Object Detection via Step-wise Learning on Smooth-tail Data}

% \author{First Author\\
% Institution1\\
% Institution1 address\\
% {\tt\small firstauthor@i1.org}
% % For a paper whose authors are all at the same institution,
% % omit the following lines up until the closing ``}''.
% % Additional authors and addresses can be added with ``\and'',
% % just like the second author.
% % To save space, use either the email address or home page, not both
% \and
% Second Author\\
% Institution2\\
% First line of institution2 address\\
% {\tt\small secondauthor@i2.org}
% }

\author{
Na Dong$^{1,2}$\thanks{Work fully done while first author is a visiting PhD student at the National University of Singapore.} \qquad Yongqiang Zhang$^2$ \qquad Mingli Ding$^2$ \qquad Gim Hee Lee$^1$ \\
$^1$Department of Computer Science, National University of Singapore\\
$^2$School of Instrument Science and Engineering, Harbin Institute of Technology\\
{\tt\small \{dongna1994, zhangyongqiang, dingml\}@hit.edu.cn \qquad gimhee.lee@comp.nus.edu.sg}
}

\maketitle
% Remove page # from the first page of camera-ready.
\ificcvfinal\thispagestyle{empty}\fi

%%%%%%%%% ABSTRACT
\begin{abstract}

Real-world data tends to follow a long-tailed distribution, where the class imbalance results in dominance of the head classes during training. 
In this paper, we propose a frustratingly simple but effective step-wise learning framework to gradually enhance the capability of the model in detecting all categories of long-tailed datasets. 
Specifically, we build smooth-tail data where the long-tailed distribution of categories decays smoothly to correct the bias towards head classes.
We pre-train a model on the whole long-tailed data to preserve discriminability between all categories.
We then fine-tune the class-agnostic modules of the pre-trained model on the head class dominant replay data to get a head class expert model with improved decision boundaries from all categories. 
Finally, we train a unified model on the tail class dominant replay data while transferring knowledge from the head class expert model to ensure accurate detection of all categories.
Extensive experiments on long-tailed datasets LVIS v0.5 and LVIS v1.0 demonstrate the superior performance of our method, where we can improve the AP with ResNet-50 backbone from 27.0\% to 30.3\% AP, and especially for the rare categories from 15.5\% to 24.9\% AP. Our best model using ResNet-101 backbone can achieve 30.7\% AP, which suppresses all existing detectors using the same backbone.
%Our source code will be made open source upon paper acceptance.  

\end{abstract}

%%%%%%%%% BODY TEXT
\vspace{-4mm}
\section{Introduction}
\label{sec:intro}

The success of deep learning are seen in many computer vision tasks including object detection. Many deep learning-based approaches~\cite{girshick2013rich,uijlings2013selective,girshick2015fast,Lin2017Feature,redmon2015you,Liu2016SSD,lin2017focal,carion2020end, zhu2020deformable} are proposed and have shown impressive performance in localizing and classifying objects of interest in 2D images. However, it is important for these deep learning-based approaches to be trained on balanced and representative datasets. Unfortunately, most real-world datasets always follow a long-tailed distribution, where the head classes have a significantly larger number of instances than the tail classes. Training on such imbalanced datasets often leads to bias towards head classes and significant performance degeneration of the tail classes due to the extremely scarce samples. 

\begin{figure}[t]
\centering
\includegraphics[width=\linewidth,height=4cm]{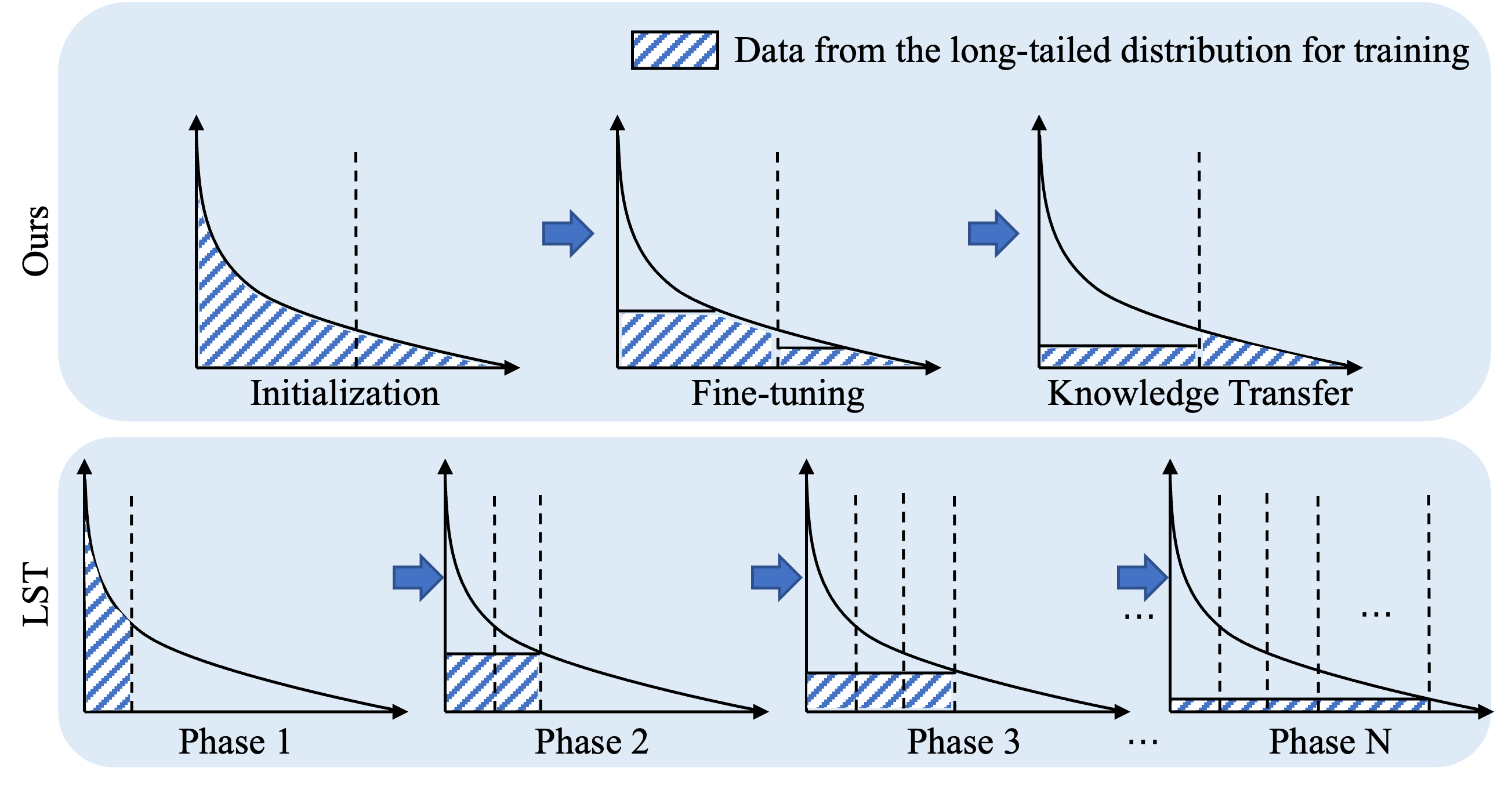}
\caption{LST~\cite{hu2020learning} is more susceptible to catastrophic forgetting due to their incremental learning scheme with numerous data splits. 
We alleviate the problem by building smooth-tail data that flattens long-tailed datasets and always maintains data from all categories. 
}
\label{fig:teaser}
\vspace{-4mm}
\end{figure}

\begin{figure}[t]
\centering
\includegraphics[width=\linewidth,height=3.9cm]{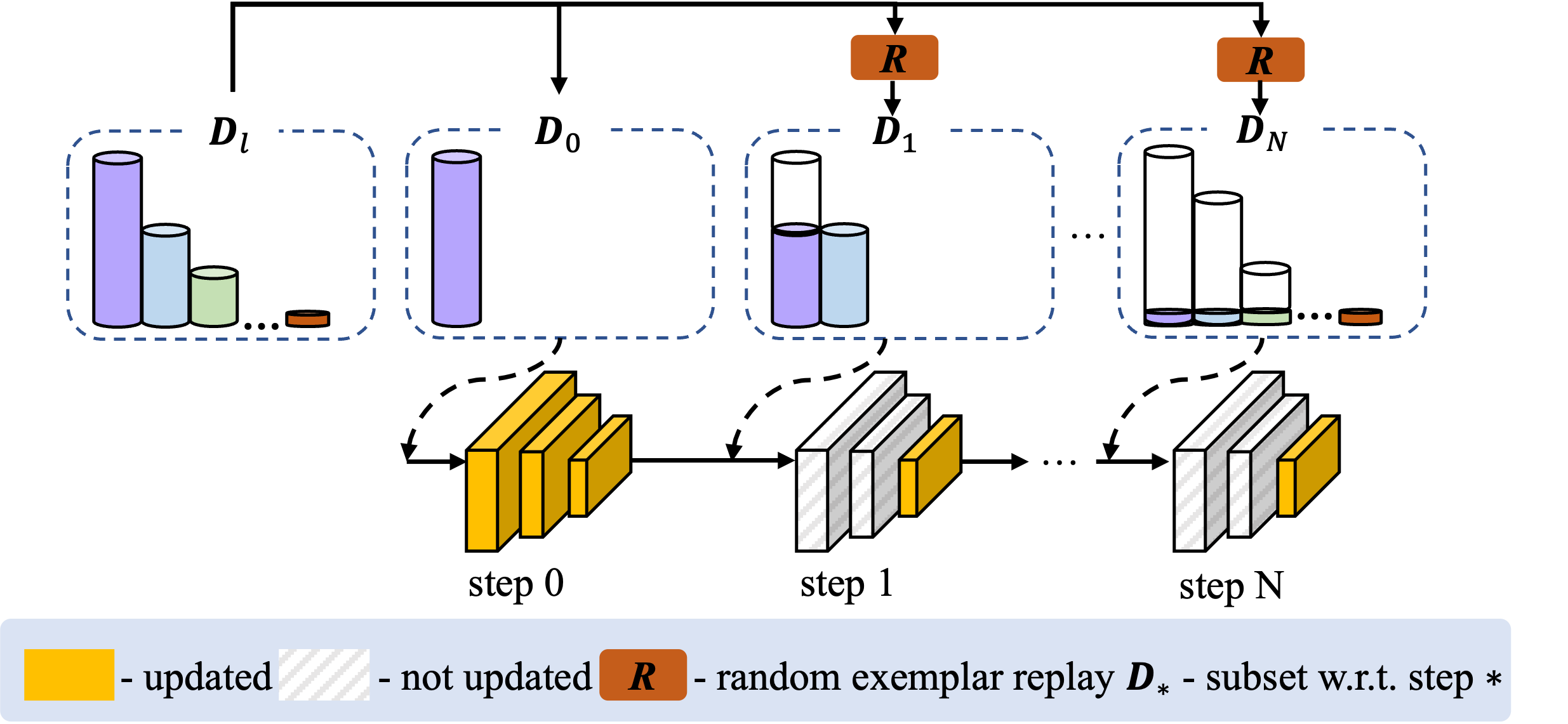}
\caption{The incremental learning training strategy of \cite{hu2020learning} on numerous smaller and balanced data splits inevitably expedites catastrophic forgetting. 
} 
\label{fig:Hu_etal}
\vspace{-4mm}
\end{figure}

To circumvent the long-tailed distribution problem of object detection task, many attempts exploit data re-sampling and loss re-weighting approaches. Data re-sampling methods~\cite{gupta2019lvis,wang2020devil} re-balance the distribution of the instance numbers of each category.
Loss re-weighting methods~\cite{tan2021equalization,wang2021seesaw,li2022equalized} adopt different re-weighting strategies to adjust the loss of different categories based on each category's statistics. 
As shown in Figure~\ref{fig:Hu_etal}, Hu \etal~\cite{hu2020learning} proposes LST which is a "divide \& conquer" strategy %which poses a learning paradigm: 
that leverages class-incremental few-shot learning to solve the long-tailed distribution problem. The model is first trained with abundant labeled data of the head classes. The categories in the long-tailed training data is then sorted and divided according to the number of samples to get the corresponding subsets for incremental learning and merging of each part in $N$ phases.
% This strategy first divides the large long-tailed dataset into parts, followed by conquering and merging each part incrementally. 
%   The categories in the long-tailed training data is then sorted and divided according to the number of samples to get the corresponding subsets for the incremental learning in N phases.

% Unfortunately, we find that this solution \etal~\cite{hu2020learning} catastrophically forgets the knowledge of head classes as well as can not sufficiently learn the tail classes in their incremental learning process. 
Despite the innovative adoption of class-incremental few-shot learning on the long-tailed distribution problem, we find that \cite{hu2020learning} catastrophically forgets the knowledge of the head classes and cannot sufficiently learn the tail classes in their incremental learning process. 
We postulate that this is attributed to three reasons: 1) Categories with high appearance similarity get divided into different parts due to the hard divisions. This leads to lower discriminability since these categories can only be trained together on the exemplar replay subsets.
% 1) Such hard division results in that categories with high appearance similarity may divided into different groups. Therefore, in subsequent process，these categories can only trained together on exemplar replay subsets, which will reduce the final model's discriminative capability between them. 
2) There is an apparent discrepancy between the decision boundaries of the current model trained simultaneously on the exemplar replay subsets of the head and tail classes from the previous model trained solely on the head class subset. 
This discrepancy impedes the maintenance of the knowledge on the head classes and the learning of the tail classes.
% 2) The model which is trained simultaneously on exemplar replay subset of  the head and tail classes, the discrepancy of decision boundary between it and the previous model which is trained solely on head classes subset is apparent.
% The discrepancy intensifies contradiction between the maintaining of the knowledge of head classes and the learning of the knowledge of tail classes, which degenerate the performance of both of them.
3) The method divides the long-tailed dataset into numerous smaller balanced parts. However, this leads to more knowledge transfer steps and thus expediting catastrophic forgetting. 
% 3) The more subsets the long-tailed dataset is divided into, the less imbalance inside each subset than the whole long-tailed dataset. However, more knowledge transfer steps between different expert model should be carried out, which causes more severe catastrophic forgetting of the knowledge of previous model and deteriorates the performance of previous categories.

In this paper, we adopt a similar incremental few-shot learning approach to the long-tailed distribution object detection problem.
%We use the state-of-the-art Deformable DETR \cite{zhu2020deformable} as our backbone object detector. 
To mitigate the above issues, we propose a simple but effective step-wise learning framework. We note that the main difference of long-tailed learning from class-incremental leaning is that the data of all categories can co-occur.
In contrast to~\cite{hu2020learning} that starts the training on only the head classes, we start the learning process from pre-training the model on the whole long-tailed dataset to better preserve the discriminative capability between the head and tail classes.
% Thus, different from~\cite{hu2020learning} that deals with the long-tailed distribution problem in a class-incremental few-shot learning way, we start the learning process from jointly-training a model on the whole long-tailed dataset as initialization, which can better preserve the discriminative capability between the head and tail classes than start from training the model only on head classes subset as ~\cite{hu2020learning}. 
In the subsequent steps, we keep the class-agnostic modules fixed and only update the class-specific modules of the pre-trained model trained on the whole long-tailed data.
%\ie, the projection layer and the classification head of Deformable DETR \cite{zhu2020deformable}.  
This circumvents the lack of training data in the tail end of the long-tailed data by preserving knowledge from the pre-trained model and limiting the network parameters that need to be updated. % the discriminative capability between all categories and original detection ability of the jointly-trained model are well kept. 

To avoid severe catastrophic forgetting, we first divide all categories of long-tailed dataset into two parts: head classes with more than $M$ images each category, and tail classes with less than $M$ images each category.
%Correspondingly
% We then propose a \N{confidence}-guided exemplar replay scheme to build:
We then propose to build smooth-tail data:
1) a head class dominant data that contain a roughly balanced subset of the head classes minored with a roughly balanced subset of tail classes, and 2) a tail class dominant data in similar vein. %for building two less imbalance but dominant by different classes subsets: 
%head class dominant subset and tail class dominant data, where the jointly-trained model is employed to select representative exemplars.
We leverage the pre-trained model to select representative exemplars for the head class dominant and tail class dominant data.
%Then, in the few-shot fine-tuning process, 
Subsequently, we fine-tune the pre-trained model on the head class dominant data to learn a head class expert model. 
Finally, we learn a unified model on the tail class dominant data while preserving knowledge of the head classes with the head class expert model. Knowledge distillation at feature level with a head class focused mask is adopt to %prevent the restraint of 
facilitate the learning of tail classes from the head class expert model. In addition, knowledge distillation at classification head is also adopted, where object query features from the head class expert model are shared to the unified model to align the predictions between them.

Our contributions can be summarized as follows: 
\begin{enumerate}[leftmargin=*] 
% \item We propose a \N{confidence}-guided exemplar replay scheme to build a head class dominant data and a tail class dominant data to prevent catastrophic forgetting in our step-wise learning framework.
\item We propose to build smooth-tail data, \ie, a head class dominant data and a tail class dominant data, to alleviate the extreme class imbalance of long-tail data
%to correct the bias towards head classes
%to remedy the performance degeneration of the tail classes
and prevent catastrophic forgetting in our step-wise learning framework.
\vspace{-2mm}
\item We design a novel step-wise learning framework that unifies fine-tuning and knowledge transfer for the long-tailed object detection task. %which is orthogonal to previous works for long-tailed object detection task.
% \item Our framework design unifies %the 
% few-shot fine-tuning and knowledge transfer methods to alleviate long-tailed distribution problem.  
\vspace{-2mm}
\item Our framework is frustratingly simple but effective. We achieve state-of-the-art performances on long-tailed datasets LVIS v0.5 and LVIS v1.0 in both the overall accuracy, and especially the impressive accuracy of the rare categories.

\end{enumerate}

\section{Related Works}
\label{sec:rela}

\paragraph{General Object Detection.}
A large number of approaches have been proposed for object detection task, which can be briefly summarized into two different types based on their frameworks.
Two-stage object detection methods such as R-CNN~\cite{girshick2013rich} apply a deep neural network to extract features from proposals generated by selective search~\cite{uijlings2013selective}. Fast R-CNN~\cite{girshick2015fast}
utilizes a differentiable RoI Pooling to improve the speed and performance. Faster R-CNN~\cite{Ren2015Faster} introduces the Region Proposal Network to generate proposals. FPN~\cite{Lin2017Feature} builds a top-down architecture with lateral connections to extract features across multiple layers.
In contrast, one-stage object detection methods such as YOLO~\cite{redmon2015you} directly perform object classification and bounding box regression on the feature maps. SSD~\cite{Liu2016SSD} uses feature pyramid with different anchor sizes to cover the possible object scales. RetinaNet~\cite{lin2017focal} proposes the focal loss to mitigate the imbalanced positive and negative examples.
Recently, %another type of 
transformer-based object detection methods~\cite{carion2020end, zhu2020deformable} beyond the one-stage and two-stage methods have gained popularity, which achieve comparable or even better performance. They directly supervise bounding box predictions end-to-end with Hungarian bipartite matching. 
These object detection models require the training datasets to possess a roughly balanced category distribution, \eg COCO dataset~\cite{lin2014microsoft}. However, the distribution of categories in the real-world scenarios is often long-tailed and most of these object detection models fail to maintain their performance. %as accurately as on the datasets with balanced category distribution, where 
An extreme imbalance leads to low accuracy on tail classes.  
% \N{To best of our knowledge, our work is the first to investigate long-tailed object detection on Deformable DETR. However, our step-wise learning framework on the smooth-tail data unifies fine-tuning and knowledge transfer, where the main components are not specific to Deformable DETR and can also be combined to other object detectors easily.}

\vspace{-2mm}
\paragraph{Long-tailed Object Detection.}
Many existing works have been proposed to alleviate the challenge of long-tailed object detection. These works can be categorized into three categories. 
\textit{Data re-sampling} is the most intuitive among all methods.
Gupta \etal~\cite{gupta2019lvis} proposes repeat factor sampling (RFS) to create a roughly balanced distribution by over-sampling data of tail classes based on the frequency of each category at image-level. Wang \etal~\cite{wang2020devil} proposes a calibration framework to alleviate classification head bias with a bi-level class balanced sampling approach at instance-level.
\textit{Loss re-weighting} is another common approach.
EQLv2~\cite{tan2021equalization} adopts a gradient-guided mechanism to re-weight the loss contribution of each category. EFL~\cite{li2022equalized} introduces a category-relevant modulating factor into focal loss to overcome the imbalance problem for one-stage object detectors. 
Wang \etal~\cite{wang2021seesaw} proposes seesaw loss to re-balance gradients of positive and negative samples for each category, with two complementary factors.
Wang \etal~\cite{wang2021adaptive} proposes to understand the long-tailed distribution in a statistic-free perspective and present a adaptive class suppression loss.
In addition to the above two common categories of methods, many works also approach the problem from different perspectives.
AHRL~\cite{li2022adaptive} addresses long-tailed object detection from a metric learning perspective, which splits the whole feature space into hierarchical structure and eliminates the problem in a coarse-to-fine manner.
Hu \etal~\cite{hu2020learning} which mainly focuses on instance segmentation task proposes to alleviate long-tailed distribution problem in a class-incremental few-shot learning way. 

\vspace{-2mm}
\paragraph{Few-Shot Object Detection and Knowledge Transfer.}
Approaches of few-shot object detection can be categorized into meta-learning based~\cite{wang2019meta,kang2019few,wu2020meta,zhang2021meta} and fine-tuning based methods~\cite{wang2020frustratingly,wu2020multi,sun2021fsce}. 
There are two key differences between few-shot object detection and long-tailed object detection.
On one hand, few-shot object detection merely focuses on the performance on few-shot categories, which is different from long-tailed object detection that aims at detecting all categories accurately.
On the other hand, the datasets of few-shot object detection are comprised of base data which contains abundant training samples per category and novel data which contains a few training samples per category, which are quite different from long-tailed datasets. 

Exemplar replay and knowledge distillation are two commonly used techniques to transfer knowledge across different models and remain performance of previous model. 
In exemplar replay based methods, the models strengthen memories learned in the past through replaying the past information periodically. They~\cite{rebuffi2017icarl, wu2019large,castro2018end} usually keep a small number of exemplars per category to achieve this purpose.
Knowledge distillation first proposed by Hinton \etal~\cite{hinton2015distilling}, where the knowledge of predicted distribution from the teacher model is distilled into the student model. Apart from the final prediction, other types of knowledge, like intermediate representations~\cite{romero2014fitnets}, can also be used to guide the learning of the student model.

Our proposed step-wise learning framework unifies fine-tuning and knowledge transfer techniques for the first time to alleviate the long-tailed distribution problem for object detection task, which can remain powerful on the head classes and better adapt to the tail classes.

\begin{figure}[t]
\centering
\includegraphics[width=\linewidth,height=2.7cm]{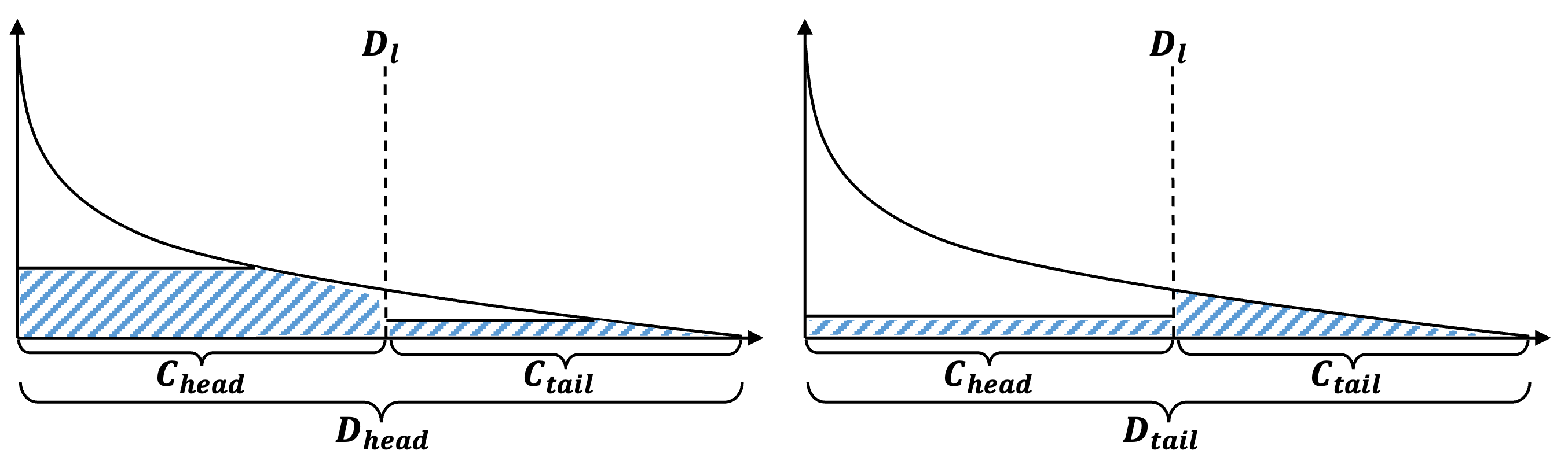}
\caption{$\mathcal{D}_{head}$ contains a roughly balanced subset of $\mathcal{C}_{head}$ and a small roughly balanced subset of $\mathcal{C}_{tail}$. $\mathcal{D}_{tail}$ contains a roughly balanced subset of $\mathcal{C}_{tail}$ and a small balanced subset of $\mathcal{C}_{head}$. 
} 
\label{fig:Datasets}
\vspace{-3mm}
\end{figure}

\begin{figure*}[!t]
\centering
\includegraphics[width=0.8\linewidth,height=4.5cm]{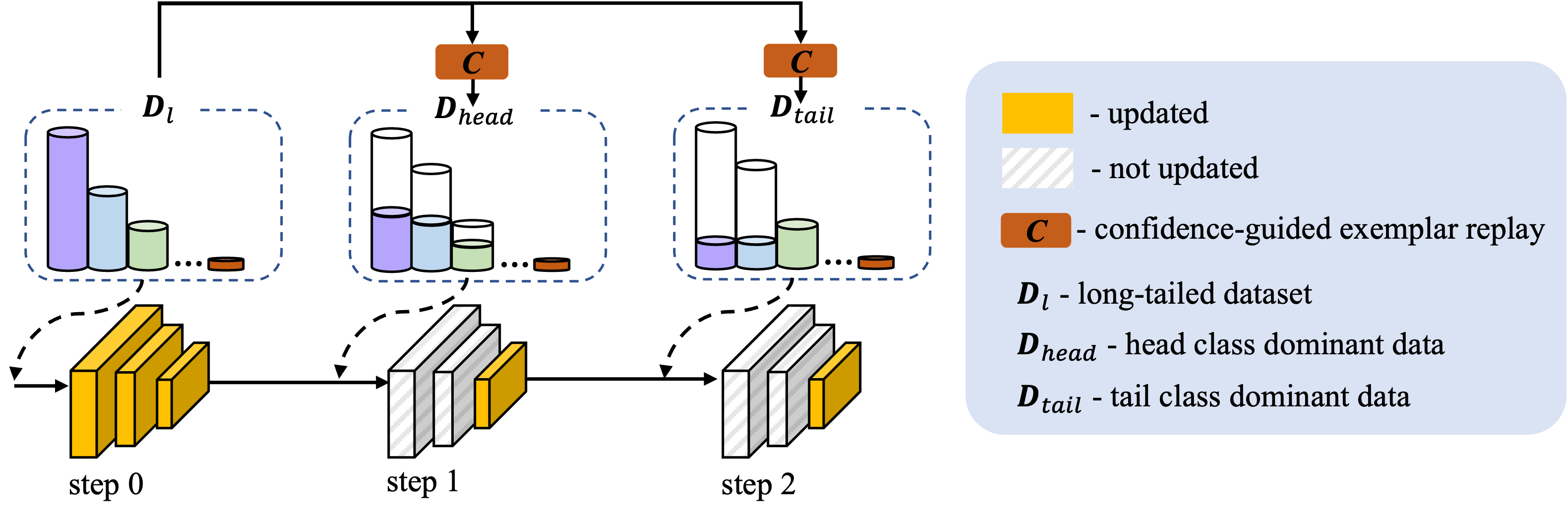}
\caption{Overview of our step-wise learning framework. We first pre-train on the whole long-tailed training data $\mathcal{D}_l$, and then the class-specific modules are fine-tuned on $\mathcal{D}_{head}$. Finally, we train the model on $\mathcal{D}_{tail}$ while concurrently preserves knowledge from $\mathcal{D}_{head}$.
} 
\label{fig:StepwiseLearning}
\vspace{-3mm}
\end{figure*}

\section{Our Methodology}
\label{sec:meth}

\subsection{Dataset Pre-processing}
As shown in Figure~\ref{fig:Datasets}, given a long-tailed dataset $\mathcal{D}_l$ with $\mathcal{C}_l$ categories, 
we divide the entire set of categories into: the head classes $\mathcal{C}_{head}$ with each category containing $\geq M$ images, and the tail classes $\mathcal{C}_{tail}$ with each category containing $<M$ images. Furthermore, $\mathcal{C}_{head} \cup \mathcal{C}_{tail} = \mathcal{C}_l$ and $\mathcal{C}_{head} \cap \mathcal{C}_{tail} = \emptyset$. We then form $\mathcal{D}_{head}$ which is dominant with a roughly balanced subset of the head classes $\mathcal{C}_{head}$ and minored with a roughly balanced subset of the tail classes $\mathcal{C}_{tail}$. Similarly, we form $\mathcal{D}_{tail}$ which is dominant with a roughly balanced subset of the tail classes  $\mathcal{C}_{tail}$ and minored with a balanced subset of the head classes $\mathcal{C}_{head}$. 

\vspace{-3mm}
% \paragraph{\N{confidence}-guided exemplar replay.}
\paragraph{Smooth-tail Data.}
%For selecting representative and diverse exemplars for the head class dominant subset $\mathcal{D}_{head}$ and the tail class dominant subset $\mathcal{D}_{tail}$, we propose a \N{confidence}-guided exemplar replay scheme. 
We propose a confidence-guided exemplar replay scheme for the selection of representative and diverse exemplars in $\mathcal{D}_{head}$ and $\mathcal{D}_{tail}$. The number of exemplars is set to be %greatly
significantly smaller than %that of 
the original dataset. 
%Aiming to ensure feasible approximation to the a valid assumption  that the model trained on the few samples also minimize its loss on the original dataset, we propose to use the jointly-trained model to %chose select the exemplars. 
We propose to use 
the model pre-trained with the whole long-tailed data (\cf next subsection) for the selection of the exemplars to ensure that the model trained on the few samples can also minimize the loss on the original dataset. Specifically, we save all instances and corresponding classification scores  $\{I_j,S_j\}$ predicted by the pre-trained model for each category. %Then, 
We then sort the instances by the value of corresponding classification scores in a descending order. Finally, we select the top-scoring instances as representative exemplars for replay. Notably, only the annotations belonging to the selected instances are considered valid in the training process. 
Furthermore, the images in original dataset are diverse in color, texture and size of region. The diversity of the exemplars ensures the same robustness and discrimination of the model as trained on original dataset, thus instances with classification scores greater than threshold $0.5$ and are not in the same image are given the priority to be chosen as exemplars.

% \subsection{Pre-trained Model} 
% Given a long-tailed dataset $\mathcal{D}_l$ with $\mathcal{C}_l$ categories, we pre-train a model on all categories using the same loss functions as Deformable DETR. 
% This pre-trained model serves to: 1) provide output classification \N{confidence}s as instance selection cues for %above 
% the \N{confidence}-guided exemplar replay scheme; 2) learn discriminative representation and provide separation capability of all categories for subsequent fine-tuning on $\mathcal{D}_{head}$ and knowledge transfer on $\mathcal{D}_{tail}$. 

\subsection{Step-wise Learning}
We use the state-of-the-art Deformable DETR \cite{zhu2020deformable} as our backbone object detector.
Given a long-tailed dataset $\mathcal{D}_l$ with $\mathcal{C}_l$ categories, we pre-train a model on all categories using the same loss functions as Deformable DETR. 
This pre-trained model serves to: 1) provide output classification confidences as instance selection cues for building the smooth-tail data; 2) learn discriminative representation and provide separation capability of all categories for subsequent fine-tuning on $\mathcal{D}_{head}$ and knowledge transfer on $\mathcal{D}_{tail}$.

%In this section, we describe the proposed step-wise learning framework in detail. 
%Given a long-tailed dataset $D_l$ with $C_l$ categories, we pre-train an initial model on all categories using the same loss functions as Deformable DETR. This initial model serves to: 1) provide output classification \N{confidence}s as instance selection cues for %above the \N{confidence}-guided exemplar replay scheme; 2) learn discriminative representation and provide separation capability of all categories for subsequent fine-tuning on $\mathcal{D}_{head}$ and knowledge transfer on on $\mathcal{D}_{tail}$. 
%Specifically, 
As shown in Figure~\ref{fig:StepwiseLearning}, we learn a head class expert model with fine-tuning, and adopt knowledge transfer from the head class expert model and the final model to unify the capability of detecting head and tail classes.
As the learning proceeds, the model gradually approaches an optimal performance of all categories.

\vspace{-3mm}
\paragraph{Fine-tuning on $\mathcal{D}_{head}$.} %In few-shot fine-tuning step, in order to learning the head classes in a balanced way, the head class dominant subset $\mathcal{D}_{head}$ which contains plenty and roughly equal amount of exemplars per head class is fed into the model. Meanwhile, the head class dominant subset $\mathcal{D}_{head}$ also contains  a small number of exemplars of the tail classes. The small number of data of tail classes is utilized to make sure the decision boundary in the feature space has smaller gap compared with the final model in subsequent step. 

We propose to only update the class-specific projection layer $\Phi_p$ and classification head $\Phi_{cls}$ with $\mathcal{D}_{head}$ while keeping the class-agnostic modules frozen. This is to impose a strong constraint on the previous representation and thus the discrimination representation does not shift severely in subsequent process.
The model is fine-tuned with the standard Deformable DETR loss~\cite{zhu2020deformable}. Note that $\mathcal{D}_{head}$ is dominant with a roughly balanced subset of $\mathcal{C}_{head}$ to alleviate class imbalance in the head classes, and minored with a roughly balanced subset of $\mathcal{C}_{tail}$ to make sure the decision boundary in the feature space has smaller gap compared to the final unified model in subsequent step.

%Given the dominating roughly balanced subset of $\mathcal{C}_{head}$ in $\mathcal{D}_{head}$ along with its annotation, 
Let the detection targets in $\mathcal{D}_{head}$ be denoted as $y = \{ y_i \}^N_{i=1} = \{(c_i, b_i)\}^N_{i=1}$, where $c_i$ and $b_i$ are the object category and bounding box. Assume the $N$ predictions for target category made by the model are $\hat{y} = \{ \hat{y}_i \}^N_{i=1} = \{  (\hat{p}(c_i), \hat{b}_i)\}^N_{i=1}$, where $\hat{p}(c_i)$ is probability of category $c_i$ and $\hat{b}_i$ is the predicted bounding box.
Following Deformable DETR, we compute the same match cost between the prediction $\hat{y}_{\hat{\sigma}(i)}$ and the ground truth  $y_i$ using Hungarian algorithm~\cite{kuhn1955hungarian}, where $\hat{\sigma}(i)$ is the index computed by the optimal bipartite matching. The Hungarian loss for all matched pairs is thus defined as:
\begin{equation}
\begin{array}{rll}
\begin{aligned}
\displaystyle 
\mathcal{L}_{hg}(y, \hat{y}) = \sum_{i=1}^N [\mathcal{L}_\text{cls}(c_i, \hat{p}_{\hat{\sigma}(i)}(c_i)) +  \mathbbm{1}_{\{c_i \neq \varnothing \}} \mathcal{L}_\text{box}(b_i, \hat{b}_{\hat{\sigma}(i)})],
\end{aligned}
\end{array}
\end{equation}
where $\mathcal{L}_\text{cls}$ is the sigmoid focal loss~\cite{lin2017focal}. $\mathcal{L}_\text{box}$ is a linear combination of $\ell_1$ loss and generalized IoU loss~\cite{rezatofighi2019generalized} with the same weight hyperparameters as Deformable DETR.

\begin{figure*}[!t]
\centering
\includegraphics[width=\linewidth,height=8cm]{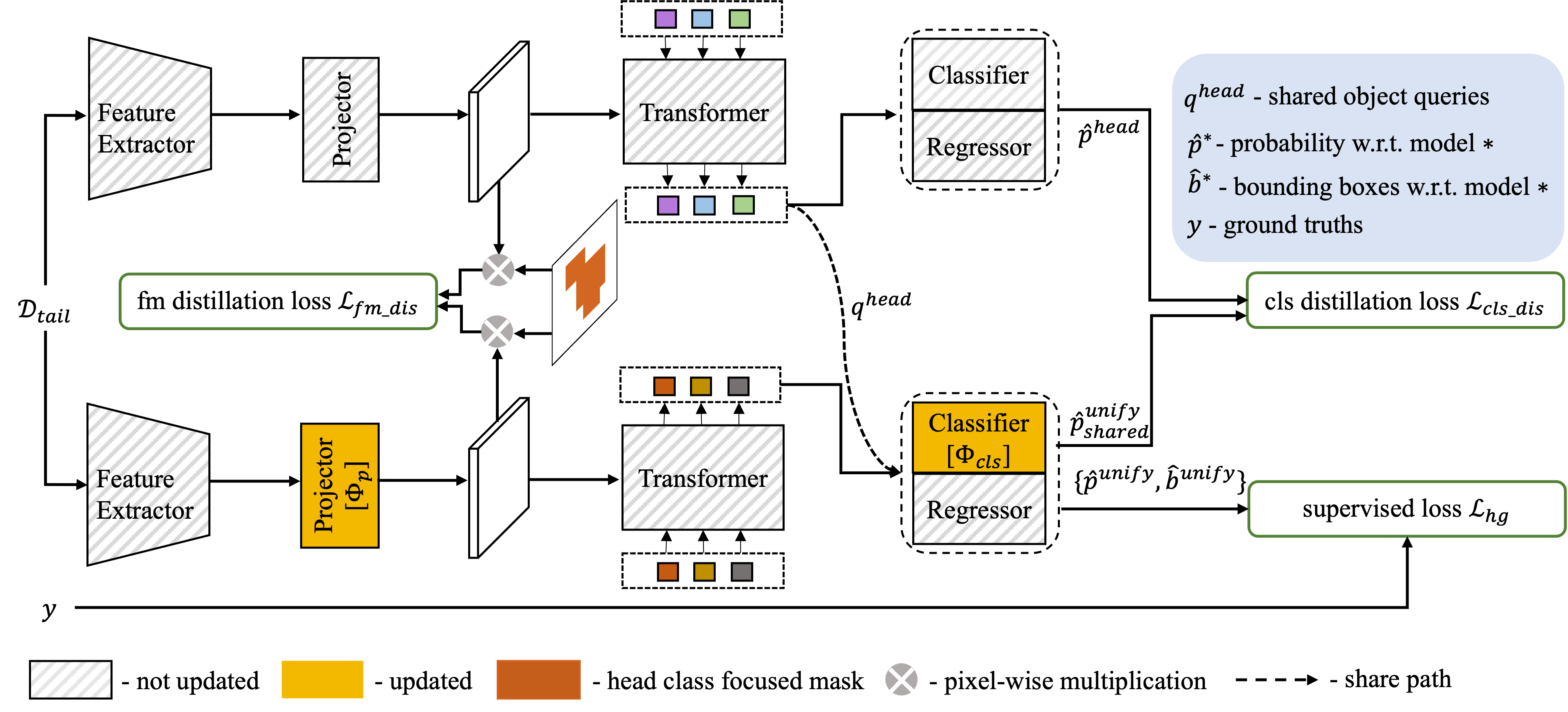}
\caption{Overview of our proposed knowledge transfer. The framework consists of the fixed head class expert model (top branch) obtained from fine-tuning on $\mathcal{D}_{head}$ for knowledge transfer to the unified model (bottom branch) during training on $\mathcal{D}_{tail}$.
} 
\label{fig2}
\vspace{-3mm}
\end{figure*}

\vspace{-3mm}
\paragraph{Knowledge Transfer on $\mathcal{D}_{tail}$.}
As shown in Figure~\ref{fig2}, we keep the model fine-tuned on $\mathcal{D}_{head}$ fixed as the head class expert model. We also keep a unified model initialized with the parameters from the head class expert model, which we train on $\mathcal{D}_{tail}$ %we initialize two models with parameters of the head class expert model obtained from the fine-tuning on $\mathcal{D}_{head}$. We keep one model frozen and train the other model on %all data of the tail classes from the long-tailed dataset to construct the tail class dominant subset as input
%$\mathcal{D}_{tail}$ 
while preserving the knowledge from $\mathcal{D}_{head}$. Similar to the fine-tuning step, we also update only the class-specific projection layer $\Phi_p$ and classification head $\Phi_{cls}$ of the unified model while keeping the class-agnostic modules frozen. However, a naive constant updates of the projection layer and classification head on the tail classes can aggravate catastrophic forgetting of the head classes. We thus propose the use of exemplar replay and knowledge distillation to mitigate the catastrophic forgetting of the head classes. 

%In the tail class dominant subset, we replay a small but equal number of exemplars per head class.
As mentioned earlier, we keep a small but balanced replay exemplars of the head classes in $\mathcal{D}_{tail}$. The head class expert model is employed as an extra supervision signal to prevent the projection layer output features of the unified model from deviating too much from the output features of the head class expert model. 
%However, in such cases,
On the other hand, we do not want the head class expert model to limit the learning process of the unified model on the tail classes. To this end, we introduce a head class focused binary mask $\mathit{mask}^\textit{head}$ based on the ground-truth bounding boxes of the head classes to prevent negative influence on the tail class learning. Specifically, we set the value of the pixel on the feature map within the ground truth bounding boxes of head classes as 1, and the value of  the pixel outside the ground truth bounding boxes as 0. The distillation loss on the features with the mask is written as:
\begin{equation}
\small
    \mathcal{L}_\textit{fm\_dis} =  \frac{1}{2N^\textit{head}} \sum_{i=1}^ w \sum_{j=1}^ h \sum_{k =1} ^c \mathit{mask}_{ij}^\textit{head} \Big\|  \mathit{f}^\textit{unify}_{ijk} - \mathit{f}^\textit{head}_{ijk} \Big\| ^2,
\end{equation}
where  $N^\textit{head} = \sum_{i=1}^w \sum_{j=1}^h \mathit{mask}_{ij}^\textit{head}$. 
$\mathit{f^\textit{head}}$ and $\mathit{f^\textit{unify}}$  denote the features of the head class expert model and the unified model, respectively. $\mathit{w}$, $\mathit{h}$ and $\mathit{c}$ are the width, height and channels of the features.

Deformable DETR is built upon the transformer encoder-decoder architecture combined with a set-based Hungarian loss that forces unique predictions for each object via bipartite matching. Object queries extract features from the feature maps. Deformable DETR learns different spatial specialization for each object query, which indicates that different object queries focus on different position areas and box sizes. 
Since there is a mismatch in the object query features input into the classification head of the head class expert model and the unified model, the predicted classification outputs between the two models can be inevitably mismatched. 
%As we directly use the prediction outputs from the head class expert model as supervisions for the unified model, since the object queries input into the classification head of each model are different, there will be mismatching between the classification outputs between them. 
% The decoder output object queries, and then these queries are inputted into detection head. But, for different model, the object queries are different, since the model selects the query match the target by Hungarian matching
To prevent the mismatch during knowledge distillation on the classification head, we first share the object query features $q^{head}$ from the decoder output of the head class expert model to align the classification probability to the unified model. %For the loss calculation, 
The classification outputs of the head class expert model and the unified model are compared in the distillation loss function given by:
\begin{equation}
\begin{array}{rll}
\begin{aligned}
\displaystyle 
\mathcal{L}_\textit{cls\_dis} &= 
 \mathcal{L}_\textit{kl\_div} ( \log (\hat{p}^\textit{unify}_\textit{shared}(c_i)), \hat{p}^\textit{head}(c_i)),  \\
\end{aligned}
\end{array}
\end{equation}
where we follow \cite{hinton2015distilling} in the definition of the KL-divergence loss $\mathcal{L}_\textit{kl\_div}$ between the category probabilities of the head class expert model and the unified model. 
$\hat{p}^\textit{unify}_\textit{shared}(c_i)$ denotes the probability of category $c_i$ with the shared object queries predicted by the unified model.
$\hat{p}^\textit{head}(c_i)$ denotes the probability of category $c_i$  predicted by the head class expert model.

A Hungarian loss $\mathcal{L}_{hg}$ is also applied to the ground truth set $y$ and the predictions $\hat{y}$ of the data of tail class dominant subset $\mathcal{D}_\textit{tail}$. The overall loss $\mathcal{L}_\textit{total}$ is given by:
\begin{equation}
\begin{array}{rll}
\begin{aligned}
\displaystyle 
\mathcal{L}_\textit{total} = \mathcal{L}_{hg}(y, \hat{y}) + \lambda_\textit{fm}\mathcal{L}_\textit{fm\_dis} + \lambda_\textit{cls}\mathcal{L}_\textit{cls\_dis}.
\end{aligned}
\end{array}
\end{equation}
$\lambda_\textit{fm}$ and $\lambda_\textit{cls}$ are hyperparameters to balance the loss terms.

\section{Experiments}
\label{sec:exper}

\begin{table*}[!htbp]
\centering
\resizebox{\linewidth}{!}{
\begin{tabular}{p{3.5cm}<{\centering}|p{2cm}<{\centering}|p{2cm}<{\centering}|p{1.5cm}<{\centering}p{1.5cm}<{\centering}p{1.5cm}<{\centering}p{1.5cm}<{\centering}}
% \begin{tabular}{ c|c|c|c|cccc}
\toprule
Method   &Backbone   & \multicolumn{1}{c|}{Dataset} & $AP^b$ & $AP_r$ & $AP_c$ & $AP_f$  \\ \hline
  LST~\cite{hu2020learning}           &\multirow{6}{*}{ResNet-50}   &  \multirow{6}{*}{LVIS v0.5}              &  22.6   & -    & -    & -    \\  
                            DropLoss~\cite{hsieh2021droploss}           &   &           &25.1    & -    & -   & -   \\  
                            EQLv2~\cite{tan2021equalization}    &    &    & 27.0    &   -  & -   &  -  \\

                        %   ACSL~\cite{wang2021adaptive}                  &   &               &-     & 18.6    & 26.4    & 29.4   \\  
                        %   LOCE~\cite{feng2021exploring}                  & &                  &28.2 & 22.0    &29.0     &30.2        \\
                           AHRL~\cite{li2022adaptive}                 &   &               &27.4     &-     & -    &-    \\
                        %   EFL~\cite{li2022equalized}                  &  &                 &     &     &     &    \\ 
                            Our  baseline               & &                  & 27.0    &  15.5   &26.9     & \textbf{31.6}   \\

                            Ours                  & &                & \textbf{30.3}          & \textbf{24.9}    &  \textbf{31.5}   & 30.9     \\
                           \hline 
                           
  LST~\cite{hu2020learning}    &\multirow{6}{*}{ResNet-101}   & \multirow{6}{*}{LVIS v0.5}           &  26.3   & -    & -    & -  \\                            
                            DropLoss~\cite{hsieh2021droploss}  &  &         &26.8    & -    & -    & -  \\ 
                            EQLv2~\cite{tan2021equalization}     &    &    & 28.1    &  -  & -    & -  \\ 
                          
                        %   Seesaw loss~\cite{wang2021seesaw}           &   &               &     &     &     &    \\  
                        %   ACSL~\cite{wang2021adaptive}                  &   &               & -    & 19.3    & 27.6    & 30.7   \\  
                        %   LOCE~\cite{feng2021exploring}                  & &                   &     &     &     &    \\
                           AHRL~\cite{li2022adaptive}                  &   &               &29.3     & -   &-     &-    \\
                        %   EFL~\cite{li2022equalized}         & CVPR 2022          &  &                 &     &     &     &    \\ 
                        Our  baseline                & &                   &27.0     & 14.6    & 27.3    & \textbf{31.7}   \\
                          
                            Ours               & &                   & \textbf{30.7}    & \textbf{26.8}    & \textbf{31.7}    & 31.1   \\
                           \hline 
 BAGS~\cite{li2020overcoming}$^\dagger$              & \multirow{7}{*}{ResNet-50}   & \multirow{7}{*}{LVIS v1.0}                    & 26.0    & 17.2    & 24.9    & 31.1  \\
%  DropLoss~\cite{hsieh2021droploss}   &  &  &22.9   &  -   & -    &  - \\ 
 EQLv2~\cite{tan2021equalization}$^\dagger$     &  &    & 25.5    & 16.4  & 23.9   & 31.2   \\ 
                           
                          Seesaw loss~\cite{wang2021seesaw}$^\dagger$           &   &               &  26.4   &  17.5   & 25.3    & 31.5   \\  
                        %   ACSL~\cite{wang2021adaptive}                 &   &               &     &     &     &    \\  
                        %   LOCE~\cite{feng2021exploring}                  & &                   & 27.4    &18.5     &26.2     &30.7    \\
                           AHRL~\cite{li2022adaptive}                  &   &               &26.4     & -    & -   &-   \\
                           EFL~\cite{li2022equalized}$^\dagger$               &  &                 & 27.5    & 20.2    & 26.1    & 32.4   \\ 
                         
                             Our  baseline                 & &                  &  25.1   &  11.9   &  23.1   & \textbf{33.2}   \\
                            Ours                  & &                   & \textbf{28.7}    & \textbf{21.8}    &\textbf{28.4}     &32.0    \\
                           \hline 
                           
 BAGS~\cite{li2020overcoming}$^\dagger$              & \multirow{7}{*}{ResNet-101}   & \multirow{7}{*}{LVIS v1.0}                   & 27.6    & 18.7    &26.5     &32.6   \\
 EQLv2~\cite{tan2021equalization}$^\dagger$       &     &   & 26.9   &  18.2  & 25.4   & 32.4   \\ 
                           
                          Seesaw loss~\cite{wang2021seesaw}$^\dagger$           &   &              & 27.8    & 18.7    & 27.0    & 32.8   \\  
                        %   ACSL~\cite{wang2021adaptive}                 &   &               &     &     &     &    \\  
                        %   LOCE~\cite{feng2021exploring}               & &                   & 29.0    &19.5     &27.8     &32.0    \\
                           AHRL~\cite{li2022adaptive}                 &   &               &28.7     & -    & -   &-   \\
                           EFL~\cite{li2022equalized}$^\dagger$                   &  &                 &29.2     & 23.5    & 27.4    &\textbf{33.8}    \\ 
                          Our  baseline          &       &                   & 26.3    & 14.4    & 24.8    & 33.2   \\
                            Ours               & &                   &  \textbf{29.5}   &\textbf{23.6}    & \textbf{29.0}    & 32.6   \\

                           \bottomrule

\end{tabular}}
\caption{\small Comparisons with the state-of-the-art methods on LVIS v0.5 and LVIS v1.0 datasets. ResNet-50 and ResNet-101 are adopted as the backbones, respectively. $^\dagger$ indicates results copied from~\cite{li2022equalized}.}
\label{comparisons}
\vspace{-2mm}
\end{table*}

\begin{table*}[!htbp] 
\centering
\resizebox{\linewidth}{!}{
\begin{tabular}{p{4cm}<{\centering}|p{3.5cm}<{\centering}|p{2cm}<{\centering}|p{2cm}<{\centering}|p{1.5cm}<{\centering}p{1.5cm}<{\centering}p{1.5cm}<{\centering}p{1.5cm}<{\centering}}
% \begin{tabular}{ c|c|c|c|cccc}
\toprule
Method   &Framework &Backbone   & \multicolumn{1}{c|}{Dataset} & $AP^b$ & $AP_r$ & $AP_c$ & $AP_f$  \\ \hline
 AHRL's baseline~\cite{li2022adaptive}   & Mask R-CNN  & \multirow{4}{*}{ResNet-50}   & \multirow{4}{*}{LVIS v0.5}   &  26.7   &  -   & -    & -   \\

                           AHRL~\cite{li2022adaptive}        & Mask R-CNN          &   &               &27.4     &-    & -    &-    \\

                         Our baseline         &Deformable DETR            &  &                 & 27.0    &  15.5   &26.9     & \textbf{31.6}   \\ 
                            Ours        & Deformable DETR          & &                &  \textbf{30.3}          & \textbf{24.9}    &  \textbf{31.5}   & 30.9    \\ \hline
EFL's baseline~\cite{li2022equalized}    & RetinaNet  & \multirow{4}{*}{ResNet-50}   & \multirow{4}{*}{LVIS v1.0}   & 25.7    &14.3     & 23.8    & 32.7   \\ 
              
                           EFL~\cite{li2022equalized}         & RetinaNet          &  &                 &  27.5   &  20.2   & 26.1    &32.4    \\ 

                         Our baseline       &Deformable DETR            &  &                 &  25.1   &  11.9   &  23.1   & \textbf{33.2}   \\ 
                            Ours        & Deformable DETR          & &                  & \textbf{28.7}    & \textbf{21.8}    &\textbf{28.4}     &32.0   \\ 
                           
                           \bottomrule
                           
% Baseline~\cite{li2022equalized}    & RetinaNet  & \multirow{4}{*}{ResNet-101}   & \multirow{4}{*}{LVIS v1.0}   & 27.0    &14.4     & 25.7    & 34.0   \\ 
              
%                           EFL~\cite{li2022equalized}         & RetinaNet          &  &                 &  29.2   &  23.5   & 27.4    &33.8    \\ 

%                          Our baseline        &Deformable DETR            &  &                 &     &     &     &    \\ 
%                             Ours        & Deformable DETR          & &                   &     &     &     &    \\ 
                           
                        %   \bottomrule

\end{tabular}}
\caption{\small Comparisons with the state-of-the-art methods and corresponding baselines.} 
\label{baseline&sota}
\vspace{-4mm}
\end{table*}

\subsection{Experimental Settings}

\paragraph{Datasets.}
To evaluate the performance of our proposed method, we conduct extensive experiments on the challenging LVIS v0.5 and LVIS v1.0 datasets. LVIS~\cite{gupta2019lvis} is a large vocabulary dataset for long-tailed visual recognition. 
LVIS v0.5 contains 1230 categories, where 57k images in the $train$ set are used for training, and 5k images in the $val$ set are used for validation. 
The latest version LVIS v1.0 contains 1203 categories, where 100k images with about 1.3M instances in the $train$ set are used for training, and 19.8k images in the $val$ set are used for validation. 
 All the categories are divided into three groups based on the number of images of each category that appear in the $train$ set: frequent (more than 100 images), common (10 to 100 images), and rare (less than 10 images). We report our results on the widely-used object detection metric $AP^b$ across IoU threshold from 0.5 to 0.95. Additionally, the boxes $AP$ for frequent ($AP_f$), common ($AP_c$), and rare ($AP_r$) categories are also reported, respectively. 

\vspace{-4mm}
\paragraph{Implementation Details.}
We implement our method on Deformable DETR~\cite{zhu2020deformable}. The ImageNet~\cite{deng2009imagenet} pre-trained ResNet-50 and ResNet-101~\cite{he2016deep} 
% with a Feature Pyramid Network~\cite{Lin2017Feature}
are adopted as the backbone. 
The training is carried out on 8 RTX 3090 GPUs with a batch size of 2 per GPU.
We train our model using the AdamW~\cite{kingma2014adam,loshchilov2017decoupled} optimizer with a weight decay of $1 \times 10^{-4}$. 
In the model pre-training step (step 0 of our framework), we train our model for 50 epochs with an initial learning rate of $2 \times 10^{-4}$ and the learning rate is decayed at 40$^\textup{{th}}$ epoch by a factor of $0.1$. 
In the model fine-tuning step (step 1 of our framework), the model is initialized from the pre-trained model. The parameters of the projection layer and classification head are updated while keeping the parameters of other modules frozen. We fine-tune the model for 1 epoch with a learning rate of $2 \times 10^{-5}$. 
In the knowledge transfer step (step 2 of our framework), the model is initialized from the fine-tuned model. The parameters of the projection layer and classification head are updated while keeping the other modules frozen. We train the model for 2 epochs with an initial learning rate of $2 \times 10^{-4}$ and the learning rate is decayed at 1$^\textup{{th}}$ epoch by a factor of $0.1$. 
$\lambda_\textit{fm}$ and $\lambda_\textit{cls}$ are set to 0.1 and 1, respectively.
% For head class dominant subset, we select 200 instances per head class. And for tail class dominant subset, select 50 instances per head class to overcome the forgetting of it.
The hyperparameter $M$ is set to 30.

\subsection{Comparisons with the State-of-the-art Methods}

To validate the effectiveness of our approach, we compare with state-of-the-art methods for long-tailed object detection on benchmark datasets LVIS v0.5 and LVIS v1.0. Our baseline is Deformable DETR~\cite{zhu2020deformable} trained on long-tailed dataset $\mathcal{D}_l$ with the same loss functions as~\cite{zhu2020deformable}.
As shown in Table~\ref{comparisons}, our method achieves the best performance compared to all other existing methods. Specifically, our proposed method achieves 30.3\% AP on LVIS v0.5 with ResNet-50 backbone. It improves the baseline by 3.3\% AP, and even achieves 9.4\% AP improvement on the rare categories. %Meanwhile, 
Our proposed method also outperforms the state-of-the-art AHRL~\cite{li2022adaptive} by 2.9\% AP. With ResNet-101 as backbone, our approach still performs well on the baseline (+3.7\% AP). %Moreover, our approach achieves superior performance to all existing methods.
Furthermore, our method outperforms the baseline by 3.6\% AP with ResNet-50 backbone and 3.2\% AP with ResNet-101 backbone on LVIS v1.0. The above results demonstrate that our method which unifies fine-tuning and knowledge transfer can effectively solve the severe class imbalance problem. 

To eliminate the doubt that whether the gain is brought by different baselines, we present a more detailed comparison with the state-of-the-art methods on both the baselines and the final models. The results are present in Table~\ref{baseline&sota}. On LVIS v0.5, our method suppresses AHRL~\cite{li2022adaptive} by 2.9\% AP with a slight advantage on baseline (AHRL's baseline: 26.7\% AP vs Our baseline: 27.0\% AP). On LVIS v1.0, while the performance of the baseline of EFL~\cite{li2022equalized} is better than our baseline (EFL's baseline: 25.7\% AP vs Our baseline: 25.1\% AP), our method still outperforms EFL~\cite{li2022equalized} by 1.2\% AP and outperforms our baseline by 3.6\% AP. %So far,
Consequently, we can conclude that the improvements brought by our method benefit from our novel design instead of the different baseline.

\subsection{Ablation Studies}
\vspace{-3mm}
% \begin{table*}[!htbp]
% \centering
% \resizebox{0.85\linewidth}{!}{
% % \begin{tabular}{ c|c|c|cccc}
% \begin{tabular}{p{3cm}<{\centering}|p{2cm}<{\centering}|p{3cm}<{\centering}|p{1cm}<{\centering}p{1cm}<{\centering}p{1cm}<{\centering}p{1cm}<{\centering}}

% \toprule
%   Baseline & Fine-tuning  & Knowledge transfer & $AP^b$ & $AP_r$ & $AP_c$ & $AP_f$  \\ \hline
    
%                       \checkmark                  &      &             & 27.0    & 15.5    & 26.9    &  31.6  \\ 
%                       \checkmark        & \checkmark        &     &  29.8                 & 20.5    &  31.5       & 31.5   \\ 
%                       \checkmark           &       & \checkmark &               29.4       &  23.2   & 29.8    &31.3    \\ 
                             
%                       \checkmark         &\checkmark     & \checkmark &            30.2          & 25.0    &  31.4   & 30.9   \\ 

%                           \bottomrule

% \end{tabular}}
% \caption{\small Ablation study of each component in our step-wise learning framework. } 
% \label{r3}
% \end{table*}

\begin{table}[!htbp]
\centering
\resizebox{\linewidth}{!}{
% \begin{tabular}{ c|c|c|cccc}
\begin{tabular}{p{1.5cm}<{\centering}|p{1.5cm}<{\centering}|p{1cm}<{\centering}p{1cm}<{\centering}p{1cm}<{\centering}p{1cm}<{\centering}}

\toprule
  FT  & KT & $AP^b$ & $AP_r$ & $AP_c$ & $AP_f$  \\ \hline
    
                            &             & 27.0    & 15.5    & 26.9    & \textbf{31.6}  \\ 
                     \checkmark        &     &  29.7                 & 19.4    &  31.4       & \textbf{31.6}   \\ 
                             & \checkmark &               29.4       &  23.2   & 29.8    &31.3    \\ 
                             
                      \checkmark     & \checkmark &           \textbf{30.3}          & \textbf{24.9}    & \textbf{31.5}   & 30.9   \\ 

                           \bottomrule

\end{tabular}}
\caption{\small Ablation study of each component in our step-wise learning framework on the smooth-tail data. FT, KT indicate the fine-tuning and knowledge transfer, respectively.} 
\label{main components}
\vspace{-4mm}
\end{table}

\vspace{-3mm}
\paragraph{Effectiveness of Each Component.} There are two steps in our proposed step-wise learning framework, \ie, fine-tuning on the head class dominant data and knowledge transfer on the tail class dominant data. We perform ablation study to demonstrate the effectiveness of each of them. As shown in Table~\ref{main components}, both the fine-tuning step and knowledge transfer step on the matched smooth-tail data play significant roles in step-wise learning framework. 

For fine-tuning the model on the head class dominant data, it improves the performance of our baseline from 27.0\% AP to 29.7\% AP, while the performance improvement on rare categories is still limited (19.4\% AP).
% We can see that fine-tuning on the head class dominant subset where exemplars for the head classes are selected by our proposed \N{confidence}-guided exemplar replay scheme can well maintain the performance on frequent categories.  Meanwhile, the improvements on common and rare categories indicates the class imbalance problem can also be alleviated in the few-shot fine-tuning way.
We then examine the effectiveness of knowledge transfer. In this setting, we directly leverage the baseline as the extra supervision in knowledge transfer step instead of using the fine-tuned head class expert model as the extra supervision. Our method outperforms the baseline by 2.4\% AP with significant improvement of the performance on the rare and common categories. However, the performance of the frequent categories experiences a slight drop.

%which can be attributed the catastrophic forgetting in the process of knowledge transfer. 
% Meanwhile, we can see the performance of rare categories is lower than that of using both of the two components. We think this is largely attribute to the baseline performs bad on tail classes and the decision boundary between the baseline and the final model is apparent, which limit the learning of the tail classes.

Fine-tuning and knowledge transfer work collaboratively to achieve an improvement from 27.0\% AP to 30.3\% AP. Particularly, it achieves 24.9\% AP for the rare categories, which outperforms the baseline by 9.4\% AP and outperforms the fine-tuned head class expert by 5.5\% AP. This indicates our proposed step-wise learning framework can sufficiently eliminate the class imbalance problem. 
However, our method experiences a further drop in the performance of the frequent categories after fine-tuning and knowledge transfer compared to using them separately (FT: 31.6\% vs KT:31.3\% vs FT\&KT: 30.9\% AP). 
% The number of images in the frequent categories is $\geq 100$, which is $\gg30$ in our head classes. 
We postulate that the drop in performance on the frequent categories might be due to insufficient representation of the frequent categories in our tail class dominant replay data during knowledge transfer.
Similarly, the selection of a roughly balanced head classes for the head class dominant replay data might also result in under representation of the frequent categories. Consequently, catastrophic forgetting has a more detrimental effect on the frequent categories.

% We think that, after fine-tuning the frequent categories on a exemplar replay set. Among the frequent categories, the performance of the categories having more data actually drops, but the performance of the categories having fewer data increases. Then, in knowledge transfer process, the performance of the categories having more data experiences further drops due to the catastrophic forgetting. 
% However, among the frequent categories, the categories having more data dominate the performance of frequent categories. 

\begin{table}[!htbp]
\centering
\resizebox{\linewidth}{!}{
% \begin{tabular}{ c|c|c|cccc}
\begin{tabular}{p{0.5cm}<{\centering}|p{1cm}<{\centering}|p{1cm}<{\centering}|p{1cm}<{\centering}p{1cm}<{\centering}p{1cm}<{\centering}p{1cm}<{\centering}}

\toprule
  SOQ  & $\mathcal{L}_\textit{fm\_dis}$ & $\mathcal{L}_\textit{cls\_dis}$ & $AP^b$ & $AP_r$ & $AP_c$ & $AP_f$  \\ \hline
    \checkmark  &  \checkmark   &                   &  29.4   & 24.9    &  30.6   & 29.8   \\
    \checkmark  &     &     \checkmark              &29.7     & \textbf{25.0}    & 30.8    & 30.3   \\
         &   \checkmark    &   \checkmark                  & 24.4    &24.3     &  26.3   & 22.0   \\
     \checkmark &   \checkmark    &   \checkmark     & \textbf{30.3}          & 24.9    &  \textbf{31.5}   & \textbf{30.9}              \\
\bottomrule
\end{tabular}}
\caption{\small Ablation study of each component in our knowledge transfer. SOQ indicates the shared object queries.} 
\label{knowledge transfer}
\vspace{-3mm}
\end{table}

\vspace{-3mm}
\paragraph{Effectiveness of Each Component of Knowledge Transfer.} We also demonstrate the effectiveness of each component of knowledge transfer. The results in  Row 1 and Row 2 of Table~\ref{knowledge transfer} show that both knowledge distillation on features and knowledge distillation on classification output predictions play significant roles in knowledge transfer. It is worth noting that the performance %dramatically 
decreases drastically when we do not share the object query features (from 30.3\% AP to 24.4\% AP), which %largely 
can be attributed to the mismatch between the classification outputs of the head class expert model and the unified model.

\begin{table*}[!t]
\centering
\resizebox{0.85\linewidth}{!}{
% \begin{tabular}{ c|cccc}
\begin{tabular}{p{9cm}<{\centering}|p{2cm}<{\centering}p{2cm}<{\centering}p{2cm}<{\centering}p{2cm}<{\centering}}
\toprule
 Division & $AP^b$ & $AP_r$ & $AP_c$ & $AP_f$  \\ \hline
                        %   $[1, -)$                    &     &     &     &    \\ 
                       
                         $[1, 10) \cup [10, -) $       &  30.1                 & 24.7    & 31.1        & 30.8   \\ 
                           $[1, 30) \cup [30, -)$  (Ours)    &\textbf{30.3}          & \textbf{24.9}    &  31.5   & 30.9       \\ 
                            $[1, 50) \cup [50, -) $       & 30.2                  &24.1    & \textbf{31.6}       & 31.0   \\ 
                         $[1, 100) \cup [100, -) $       &  30.1                 & 23.9    &  31.3      & \textbf{31.2}   \\ 
                        $[1,10) \cup [10, 100) \cup [100, -)$     & 29.8                     & 23.7    &31.0     & 30.7   \\ 
                         $[1,10) \cup [10, 30) \cup [30, 100) \cup [100, -)$     &  29.3                    &  24.8   & 30.4    & 29.8   \\ 

                           \bottomrule
\end{tabular}}
\caption{\small Ablation study of different type of divisions.} 
\label{divisions}
\vspace{-4mm}
\end{table*}

\vspace{-3mm}
\paragraph{Analysis of Divisions.}
The type of divisions on the long-tailed data plays an important role in our approach. We conduct extensive experiments to study the influence of different type of divisions of the long-tailed dataset. As shown in Table~\ref{divisions}, we can see that training the model with division $[1, 30) \cup [30, -)$ achieves the best performance. All two-step divisions can outperform the performance of three-step or four-step divisions. We attribute this good performance to the fewer divisions, and the lower performance by the divisions $[1,10) \cup [10, 100) \cup [100, -)$ and  $[1,10) \cup [10, 30) \cup [30, 100) \cup [100, -)$ are caused by severe catastrophic forgetting %brought by more steps
from the increase in divisions. The performance of our two-step division $[1, 30) \cup [30, -)$ also surpasses the other three two-step divisions, which clearly demonstrate the superiority of the division $[1, 30) \cup [30, -)$ in adapting to the tail classes while maintaining the performance of the head classes.

\begin{table}[!t]
\centering
\resizebox{\linewidth}{!}{
% \begin{tabular}{ c|cccc}
\begin{tabular}{p{2cm}<{\centering}|p{2cm}<{\centering}|p{1cm}<{\centering}p{1cm}<{\centering}p{1cm}<{\centering}p{1cm}<{\centering}}
\toprule
 $N_{ex}$ of $\mathcal{C}_{head}$  &$N_{ex}$ of $\mathcal{C}_{tail}$ & $AP^b$ & $AP_r$ & $AP_c$ & $AP_f$  \\ \hline
                       
                         100     &30   & 30.1         & \textbf{25.0} &   31.3 & 30.6   \\ 
                           200    &30     &\textbf{30.3}          & 24.9    & \textbf{31.5}   & 30.9    \\ 
                            300    &30  & 30.0         &24.6  & 31.0   & \textbf{31.0}    \\ \hline
                        %  200     &0   &          &  &    &    \\ 
                        200     &10   & 30.2         &24.7  & 31.3   & \textbf{30.9}   \\ 
                           200    &30    &\textbf{30.3}          & 24.9    &  \textbf{31.5}   & \textbf{30.9}    \\        
                           200    &50  & 30.2         &25.0  & 31.3   & 30.8       \\ 
                            200    &100  & 30.2         & \textbf{25.1} &\textbf{31.5}    & 30.8    \\

                           \bottomrule
\end{tabular}}
\caption{\small Ablation study of exemplar memory size of $\mathcal{D}_{head}$.} 
\label{size1}
\vspace{-3mm}
\end{table}

\begin{table}[!t]
\centering
\resizebox{\linewidth}{!}{
% \begin{tabular}{ c|cccc}
\begin{tabular}{p{4cm}<{\centering}|p{1cm}<{\centering}p{1cm}<{\centering}p{1cm}<{\centering}p{1cm}<{\centering}}
\toprule
 $N_{ex}$ of $\mathcal{C}_{head}$  & $AP^b$ & $AP_r$ & $AP_c$ & $AP_f$  \\ \hline
                        %  0     &          &  &    &    \\ 
                         10     & 29.6         &24.4  & 30.8 &30.2       \\ 
                           30    &  30.0        &24.8  & 31.2   & 30.6       \\ 
                            50        &\textbf{30.3}          & \textbf{24.9}   & \textbf{31.5}   & 30.9    \\ 
            
                             100     & 30.1         &23.2  &31.4    & \textbf{31.3}    \\

                           \bottomrule
\end{tabular}}
\caption{\small Ablation study of exemplar memory size of $\mathcal{D}_{tail}$.} 
\label{size2}
\vspace{-4mm}
\end{table}

\vspace{-3mm}
\paragraph{Analysis of Exemplar Memory Size.} 
We form $\mathcal{D}_{head}$ which is dominant with a roughly balanced subset of the head classes $\mathcal{C}_{head}$ and minored with a roughly balanced subset of the tail classes $\mathcal{C}_{tail}$. Similarly, we form $\mathcal{D}_{tail}$ which is dominant with a roughly balanced subset of the tail classes  $\mathcal{C}_{tail}$ and minored with a balanced subset of the head classes $\mathcal{C}_{head}$. 
% We propose a \N{confidence}-guided exemplar replay scheme for the selection of representative and diverse exemplars in $\mathcal{D}_{head}$ and $\mathcal{D}_{tail}$.
 We denote $N_{ex}$ as the number of instances per category. For $\mathcal{D}_{head}$ and $\mathcal{D}_{tail}$, we vary $N_{ex}$ of the head classes $\mathcal{C}_{head}$ and the tail classes $\mathcal{C}_{tail}$ and report the results in Tables~\ref{size1} and~\ref{size2}, respectively.
We find that increasing $N_{ex}$ of $\mathcal{C}_{head}$ helps maintain the performance of head classes. However, we also observe that increasing $N_{ex}$ of $\mathcal{C}_{head}$ impedes the learning of tail classes and hurts the performance of tail classes. 
In addition, increasing $N_{ex}$ of $\mathcal{C}_{tail}$ to large values does not significantly help the learning of the tail classes and slightly shows adverse affects on the performance of the head classes.
By validation, we therefore store 200 instances per category of $\mathcal{C}_{head}$ and 30 instances per category of $\mathcal{C}_{tail}$ in $\mathcal{D}_{head}$. Similarly, in $\mathcal{D}_{tail}$, we store 50 instances per category of $\mathcal{C}_{head}$ and introduce all instances of $\mathcal{C}_{tail}$. 
This can eliminate the class imbalance between $\mathcal{C}_{head}$ and $\mathcal{C}_{tail}$ inside the exemplar sets and achieve a trade-off of the performance of all categories.

\begin{table}[!t]
\centering
\resizebox{\linewidth}{!}{
% \begin{tabular}{ c|cccc}
\begin{tabular}{p{4cm}<{\centering}|p{1cm}<{\centering}p{1cm}<{\centering}p{1cm}<{\centering}p{1cm}<{\centering}}
\toprule
 Method & $AP^b$ & $AP_r$ & $AP_c$ & $AP_f$  \\ \hline
                           Ours w/o step-wise RFS                    & 29.6    &  19.0   & \textbf{31.6}    & \textbf{31.2}   \\ 
                       
                            Ours  &\textbf{30.3}          & \textbf{24.9}    &  31.5   & 30.9    \\ 
             
                           \bottomrule
\end{tabular}}
\caption{\small Ablation study of step-wise RFS.} 
\label{RFS}
\vspace{-4mm}
\end{table}

\vspace{-3mm}
\paragraph{Analysis of Step-wise RFS.}
Class imbalance still exists in the exemplar replay data for the head and tail classes due to the severe imbalance between categories of the long-tailed dataset, and thus hinders the learning of categories having fewer data. 
% Due to the severe imbalance between categories of the long-tailed dataset and the extremely scarce samples of the tail classes, the imbalance inside the dominant classes of the two subsets still exists. $\mathcal{D}_{head}$ and $\mathcal{D}_{tail}$
To narrow the imbalance in the exemplar replay data, we propose to adopt the repeat factor sampling (RFS) to over-sample the data from categories having fewer data. In our proposed step-wise learning framework, 
% RFS is used with different hyperparameters in different steps and thus we terms it as step-wise RFS.
RFS is used in different ways in different steps and thus we terms it as step-wise RFS.
In the fine-tuning step, for the head class dominant replay data, we over-sample the categories having fewer data among the dominant head classes. In the knowledge transfer step, we also over-sample the categories having few data among the dominant tail classes for the tail class dominant replay data. As shown in Table~\ref{RFS}, the comparisons between our method using and without using step-wise RFS indicate that applying step-wise RFS does help alleviate the imbalance inside the subsets.

\section{Conclusion}
\label{sec:conc}
\vspace{-2mm}
In this work, we propose a simple yet effective method 
%from an orthogonal perspective from previous works to address 
that leverages incremental learning on the long-tailed distribution problem for the object detection task. We identify that a pre-trained model on the whole long-tailed dataset can achieve high discriminability in all categories for subsequent training steps.
%Meanwhile, 
% A \N{confidence}-guided exemplar replay scheme is introduced  for the selection
% of representative and diverse head and tail class exemplar replay data.
We propose to build the smooth-tail distributed data for calibrating the class imbalance in long-tailed datasets, and maintaining representative and diverse head and tail class exemplar replay data.
We propose a novel step-wise learning framework that first fine-tune the pre-trained model on the head class dominant replay data to get the head class expert model. Subsequently, knowledge is transferred from the head class expert model to a unified model trained on the tail class dominant replay data. %It transfer knowledge across different classes expert models and obtain a unified model have optimal performance for all categories finally. 
Our method brings large improvements with notable boost on the tail classes on different backbones and various long-tailed datasets. Furthermore, our method achieves state-of-the-art performance on the challenging LVIS benchmarks for object detection task.

% Th performance drop in head classes still exists.

\clearpage

% {\small
% \bibliographystyle{ieee_fullname}
% \bibliography{egbib}
% }

\end{document}